%% file: old_technical_paper.tex
\documentclass[11pt,a4paper]{article}

\usepackage[utf8]{inputenc} 
\usepackage[T1]{fontenc}    
\usepackage{graphicx}       
\usepackage{amsmath}        
\usepackage{url}            
\usepackage[colorlinks=true, linkcolor=blue, citecolor=blue, urlcolor=blue]{hyperref} 
\usepackage[blocks]{authblk} 
\usepackage{booktabs}       
\usepackage[authoryear]{natbib}         
\usepackage[margin=1in]{geometry} 
\usepackage{enumitem} 
\usepackage{caption}        
\usepackage{longtable}      
\usepackage{parskip}        
\usepackage{framed}         
\usepackage[most]{tcolorbox}
\usepackage{hyperref} 
\usepackage{xurl}     

\usepackage{xcolor}

\hypersetup{
    colorlinks=true,   
    urlcolor=blue      
}

\tcbset{
  mygreybox/.style={
    colback=gray!10,
    colframe=gray!50,
    boxrule=0.5pt,
    arc=2mm,
    left=2mm,
    right=2mm,
    top=1mm,
    bottom=1mm,
    enhanced,
    breakable    
  }
}

\title{AI Agents for Conversational Patient Triage: Preliminary Simulation-Based Evaluation with Real-World EHR Data}

\author[1]{Sina Rashidian PhD}
\author[1]{Nan Li PhD}
\author[1]{Jonathan Amar PhD}
\author[1]{Jong Ha Lee M.S.}
\author[1]{Sam Pugh M.S.}
\author[1]{Eric Yang MBI}
\author[1]{Geoff Masterson MBA}
\author[1]{Myoung Cha JD, MBA}
\author[1]{Yugang Jia MPH PhD\thanks{Corresponding author: Yugang Jia, PhD\newline Verily Life Sciences\newline 2999 Olympus Blvd, Ste 1000,\newline Dallas, TX 75019\newline P. 650-495-7100\newline E. \href{mailto:yugang@verily.com}{yugang@verily.com}}}
\author[2,3]{Akhil Vaid MD}

\affil[1]{Verily Life Sciences, 999 Bayhill Drive, San Bruno, CA}
\affil[2]{Windreich Department for AI and Human Health, Icahn School of Medicine at Mount Sinai, New York, NY}
\affil[3]{Division of Data Driven and Digital Medicine, Icahn School of Medicine at Mount Sinai, New York, NY}

\date{} 

\begin{document}




\maketitle 

\vspace{1cm}



\begin{framed}
\noindent\textbf{Key points}
 are summarized as follows:
\begin{itemize}
    \item A novel approach to derive patient vignettes from real world EHR data that powers a patient simulator and generate scalable conversational encounters for AI triage agent development. 
    \item A multi-agent architecture that conducts comprehensive conversational patient assessment, on-demand EHR data retrieval, iterative reasoning for differential diagnosis, and the generation of clinical guidelines-based triaging advice.
    \item Preliminary results of 519 simulated encounters evaluated by 2 experienced clinicians across 14 dimensions including synthetic patient vignette quality, information gathering, clinical reasoning, and care recommendations.

\end{itemize}
\end{framed}

\clearpage 

\begin{abstract}
\noindent \textbf{Background:}
Since the release of GPT 3.5 (ChatGPT) in November 2022, the Overton window has shifted for the utility of large language models (LLMs) in almost all aspects of life.  Executives use these tools as a thought-companion; teachers and students alike have found ways to integrate these tools to personal education; physicians and patients have each individually explored how LLMs can help with diagnosis, improve patient and physician experience, and even supporting critical medical decision making. In less than three years, the performance of LLMs has surpassed human ability in many tasks of human performance, such as medical exams or mathematic olympiads, but to have the greatest impact in the clinical domain, LLMs will need to demonstrate performance in trusted multi-turn conversation - the analog of building a trusted patient-doctor relationship.  

In this work, we present a Patient Simulator that leverages real world patient encounters which cover a broad range of conditions and symptoms to provide synthetic test subjects for development and testing of healthcare agentic models.  Using real patient interactions introduces legal and ethical barriers and has been a major barrier to development of scalable personalized healthcare agents.  The simulator overcomes these barriers by providing a realistic approach to patient presentation and multi-turn conversation with a symptom-checking agent, all related to real case studies taken from abstracts of patient encounters. 

\noindent \textbf{Objectives:}
(1) To construct and instantiate a Patient Simulator to train and test  an AI health agent, based on patient vignettes derived from real EHR data.
(2) To test the validity and alignment of the simulated encounters provided by the Patient Simulator to expert human clinical providers. 
(3) To illustrate the evaluation framework of such an LLM system on the generated realistic, data-driven simulations -- yielding a preliminary assessment of our proposed system.

\noindent \textbf{Methods:}
We first constructed realistic clinical scenarios by deriving patient vignettes from real-world EHR encounters. These vignettes cover a variety of presenting symptoms and underlying conditions common in the primary care setting. 
We then evaluate the performance of the Patient Simulator as a simulacrum of a real patient encounter across over 500 different patient vignettes.  We evaluated the efficacy of retrieval from the patient vignettes.  We leveraged a separate AI agent to provide multi-turn question to obtain a history of present illness. The resulting multiturn conversations were evaluated by two expert clinicians as to whether the responses were appropriate given the clinical vignette using structured questionnaires and free form responses.

\noindent \textbf{Results:} 
In the multi-turn conversations with a symptom-checking AI agent, clinicians scored the Patient Simulator as consistent with the patient vignettes in those same 97.7\% of cases and across the variety of disease states represented in the vignettes. The extracted case summary based on the conversation history was 99\% relevant.



\noindent \textbf{Conclusions:}
We developed a methodology to incorporate vignettes derived from real healthcare patient data to build a simulation of patient responses to symptom checking agents.  The performance and alignment of this Patient Simulator could be used to train and test a multi-turn conversational AI agent at scale. We demonstrate the Patient Simulator's consistency with the real patient encounter, by having two clinicians independently evaluate the simulated conversations in the context of each vignette. The Patient Simulator represents a powerful tool to develop and evaluate realistic and acceptable agentic systems for healthcare settings without risking patient privacy. 


\end{abstract}


\clearpage 

\section{Introduction}
Timely triage, diagnosis, and medical decisions are key in limiting morbidity and improving overall patient outcomes \citep{Iserson2007}. However, healthcare access gaps persist due to factors such as geographic limitations or resource constraints; and addressing this need becomes critical as many of the leading contributors to global disease burden are conditions that could be mitigated through timely and equitable care access \citep{Bonow2005, Chandran2022, Capps2003, Murray2024}. Yet, while overall access to clinical care is a multifactorial issue, individual access to relevant medical information early-on in the care process is a universal key component, and often an inefficient step in the overall patient pathway.

The integration of artificial intelligence (AI)-driven solutions, particularly generative large language models \citep{Thirunavukarasu2023, Mielke2021} (LLMs), in medical care has emerged as a promising approach to addressing information access gaps and delays. It is possible to implement LLM-driven agents \citep{Huang2024}, i.e. models operating within language enforced constraints and which, given an appropriate set of operating instructions, can operate autonomously and navigate complex environments. Agents are multi-purpose. They do not require specialized training for each new condition and very importantly use natural language \citep{Chowdhary2020} as their primary interface.
The application of LLM-based agents to the delivery of clinical information represents a leap in functionality over traditional modeling approaches relying on Electronic Healthcare Record (EHR) derived data \citep{Goldstein2017, Rasmy2021, Li2020, Yang2022}. They can engage in dynamic, context-aware, adaptable and broad-spectrum conversations with patients \citep{Huang2024} without the need for extensive customization \citep{Vaid2023} or predefined structured inputs \citep{Huang2024, Vaid2023, Feder1999}, unlike static clinical decision support tools. By integrating with mobile health platforms, LLM-driven agents could also assist in remote patient monitoring and consultation or personalized patient education \citep{moritz2025coordinated}.

While promising, current AI-driven approaches in healthcare face significant limitations. Existing systems \citep{Schmidgall2024b, Tu2024}, often relying on single-agent architectures, may not fully leverage advanced LLM reasoning. Furthermore, their decision-making processes can be opaque, a critical issue stemming from the black-box, stochastic nature of LLMs, which hinders interpretability, reproducibility, and practitioner trust \citep{Li2024, Naveed2024}. Although efforts like integrating web search tools enhance specific functionalities like question-answering accuracy \citep{Zakka2024, Tu2024, Nakano2021}, the underlying challenges of opacity and verification persist. Multi-agent systems, employing well-designed collaborative architectures, represent a potential solution, offering enhanced transparency, control, and mechanisms for step-by-step verification compared to single-agent models. Nevertheless, the evidence base for integrating these multi-agent systems effectively into clinical workflows remains limited.

A critical challenge in deploying LLM-driven agent systems in clinical settings is how to rigorously test and debug them prior to launch. A thorough evaluation requires access to real patient data and the ability to simulate dynamic, evolving clinical scenarios grounded in realistic patient profiles.
Extant attempts have made stringent assumptions by generating fully synthetic patients with auxiliary LLMs \citep{Tu2024, Schmidgall2024b}, limiting the generalizability of the evaluation procedure, without grounding on real patient data.
Our approach, on the other hand, leverages the depth of real-world electronic health record (EHR) data to build such simulations at scale across various conditions, allowing meaningful assessment of clinical accuracy and agent reasoning.
Our simulations utilize comprehensive patient data, including historical labs and notes, to capture the specificity of individual real-world encounter rather than the simplifying examples directly generated by an auxiliary LLM.
Beyond testing, an LLM-system in a clinical setting must also be debuggable and extensible. A single LLM model lacks transparency and control, making it difficult to isolate errors or trace decision logic. By contrast, a modular, multi-agent architecture structured around physician workflows and clinical reasoning steps which offers clearer traceability, interpretability, and the flexibility to iteratively refine and expand system capabilities.

In this report, we propose a Patient Simulator that enables the scalable testing of AI agents using well-designed evaluation criteria \citep{lizee2024conversational}. The Patient Simulator dynamically adapts real-world patient encounters into conversations with an AI agent.
Then we present our proposed approach for triaging, AI Triage: a multi-agent system composed of eight LLM-based agents designed to emulate the clinical reasoning process of a physician. The system mirrors the typical diagnostic workflow starting from SOCRATES-based symptom collection, progressing through patient summarization, EHR retrieval, differential diagnosis (DDX), care recommendation, and guideline verification to arrive at a triage decision.
In preliminary assessments against our proposed Patient Simulator, the multi-agent system engaged in simulated clinical conversations with the Patient Simulator, gathering relevant symptom information, incorporating available clinical data, and issuing triage recommendations. Two human experts reviewed these outputs and expressed agreement with the triage decisions and rationale, and positively reviewed related system functions through structured questionnaires and free text.

\section{Methodology}
\begin{figure}[htbp]
    \centering
    \includegraphics[width=1\textwidth]{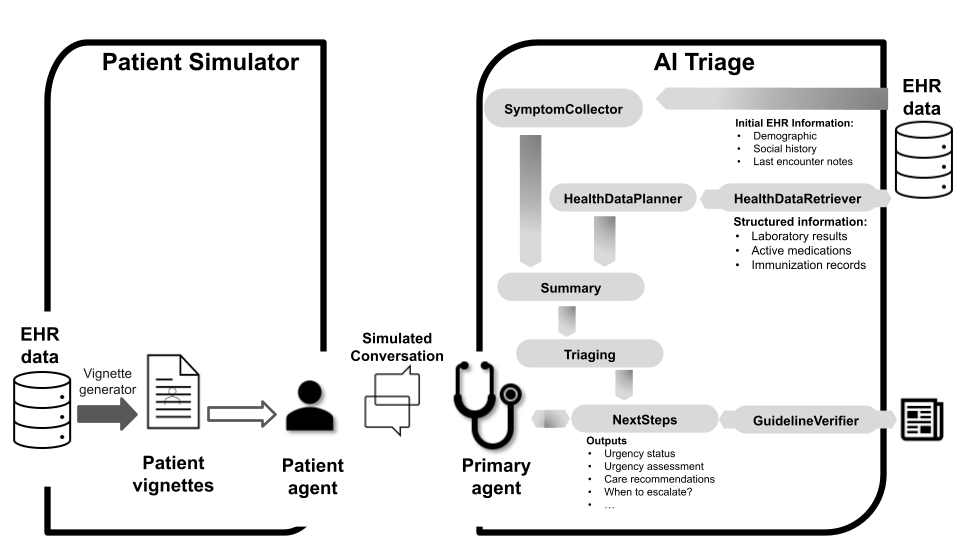} 
    \caption{Schematic description of the Patient Simulator, the AI Triage agent, and their interaction.}
    \label{fig:schematic}
\end{figure}

We describe the Patient Simulator, which is the main contribution to evaluating AI systems by extending EHR data into conversations for downstream review. Then we detail our proposed architecture for AI Triage, a system investigating patient symptoms driven by a debuggable and interpretable multi-agent system.
The interaction between these two contributions is represented in Figure \ref{fig:schematic}.


\subsection{Patient Simulator}
The Patient Simulator is the backbone for evaluating a triage system at scale across diverse conditions. The system is built on top of real-world clinical encounters to instantiate conversations for diverse patient behavior with various symptoms.

The simulator then acts as a responsive client to the AI Triage and is designed to provide knowledge of its condition while restricting the disclosure of information beyond what could be elicited through conversation. To construct the Patient Simulator, details from each single clinical encounter were transformed into a single patient vignette, serving as the foundation for subsequent simulated interactions, Figure~\ref{fig:schematic} left. The Patient Simulator was instructed to respond based on the details contained within the vignette, generating plausible inferences when necessary while maintaining internal consistency. Additionally, to mirror real-world patient interactions, the Patient Simulator described symptoms in natural language, avoiding jargon and clinical terminology. These constraints ensured that the simulated patient responses remained contextually grounded and reflective of real-world clinical encounters.

The desired behavior of the Patient Simulator was iteratively improved with clinical expert guidance, resulting in a set of instructions to follow. The instructions are meant to be clear and concise with key examples illustrating desired behavior.
For details, see the Patient Simulator prompt in Appendix \ref{sec:supp_prompt}.

\subsection{AI Triage Architecture}

The AI Triage system operated through a multi-agent workflow \citep{Talebirad2023}, powered by multiple LLMs, to represent the different components of a physician's decision-making process (Figure~\ref{fig:schematic} right). The desirable attributes for the system as a whole were to be able to manage real-time, coordinated interaction between multiple AI agents for comprehensive patient assessment (based on simulation, for the purpose of this study). These agents engaged in a seamless interactive dialogue, simulating the diagnostic reasoning and information-gathering processes that occur in real-world clinical encounters. The system was expected to conduct a realistic clinical visit interaction, with the individual AI agents collectively extracting critical details, mirroring the way human physicians dynamically adjust their approach based on evolving patient information and data.
We provide an overview of the different agents involved in AI Triage in Table \ref{tab:ai-triage-main}, describing objectives, input, outputs and key properties.
From the patient's perspective, the multi-agent system interacts as one conversational entity, as would an interaction with a single care contact.

\footnotesize
\begin{longtable}{l|l} 
\caption{Overview of AI Triage internal agents with key features}
\label{tab:ai-triage-main}
\\

\toprule 
\textbf{Agent name} & \textbf{Description}\\
\midrule 
\endhead

\midrule
Primary
& \textbf{Objective:} Orchestration, fixing tone and format\\
& \textbf{Multi-turn:} Yes\\

\midrule
SymptomCollector
& \textbf{Objective:} Collecting symptoms\\
& \textbf{Input:} User responses + Initial EHR data \\
& \textbf{Output:} Conversation containing chief complaint and symptoms\\
& \textbf{Multi-turn:} Yes\\
& \textbf{Notes:} Uses RAG to digest patient level EHR beginning of the conversation\\

\midrule
HealthDataPlanner
& \textbf{Objective:} Devise which health data would be relevant in reasoning\\
& \textbf{Input:} SymptomCollector's chat history\\
& \textbf{Output:} List of needed EHR data to narrow down the triage and ddx.\\
& \textbf{Multi-turn:} Yes - confirms data is the most recent\\

\midrule
HealthDataRetriever
& \textbf{Objective:} Retrieve health data from EHR leveraging plan\\
& \textbf{Input:} Data Plan from HealthDataPlanner\\
& \textbf{Output:} EHR records retrieved from database.\\
& \textbf{Multi-turn:} No\\

\midrule
Summary
& \textbf{Objective:} Synthesize gathered information into initial case summary\\
& \textbf{Input:} SymptomCollector's conversation + retrieved data from HealthDataPlanner\\
& \textbf{Output:} Semi-structured case summary\\
& \textbf{Multi-turn:} No\\

\midrule
Differential Diagnosis
& \textbf{Objective:} Narrowing down differential diagnosis\\
& \textbf{Input:} Case summary + User responses\\
& \textbf{Output:} Questions to narrow down ddx\\
& \textbf{Multi-turn:} Yes\\
& \textbf{Notes:} Internal step-by-step reasoning, how each question narrows down the ddx. \\

\midrule
Next Steps
& \textbf{Objective:} Provide recommendations\\
& \textbf{Input:} Comprehensive case summary + differential diagnosis chat history\\
& \textbf{Output:} Assessments, urgency status, care instructions and when to escalate\\
& \textbf{Multi-turn:} No\\

\end{longtable}
\normalsize

To enhance the framework’s utility, the Next Steps Agent provided tailored recommendations for home care when self-care and follow-up were deemed appropriate. These instructions included symptom management strategies, lifestyle modifications, and guidance on medication adherence when applicable. Additionally, the system generated a list of warning signs specific to the patient’s condition. If any of these indicators emerged, the patient was advised to seek a higher level of care, ensuring timely medical intervention when necessary. This approach aimed at empowering patients with actionable insights while maintaining a safety net for escalating care when required.

The main output of the system as a whole was intended to be a contextually relevant and personalized triage recommendation, formed through dynamic integration across various data types (including, but not limited to, patient-reported information such as symptoms, i.e. labs, medications) and complex step-by-step reasoning capabilities, which sheds light into the decision making process, giving rise to enhanced interpretability and transparency.

\subsection{Guideline Verifier}

\begin{figure}
    \centering
    \includegraphics[width=1\linewidth]{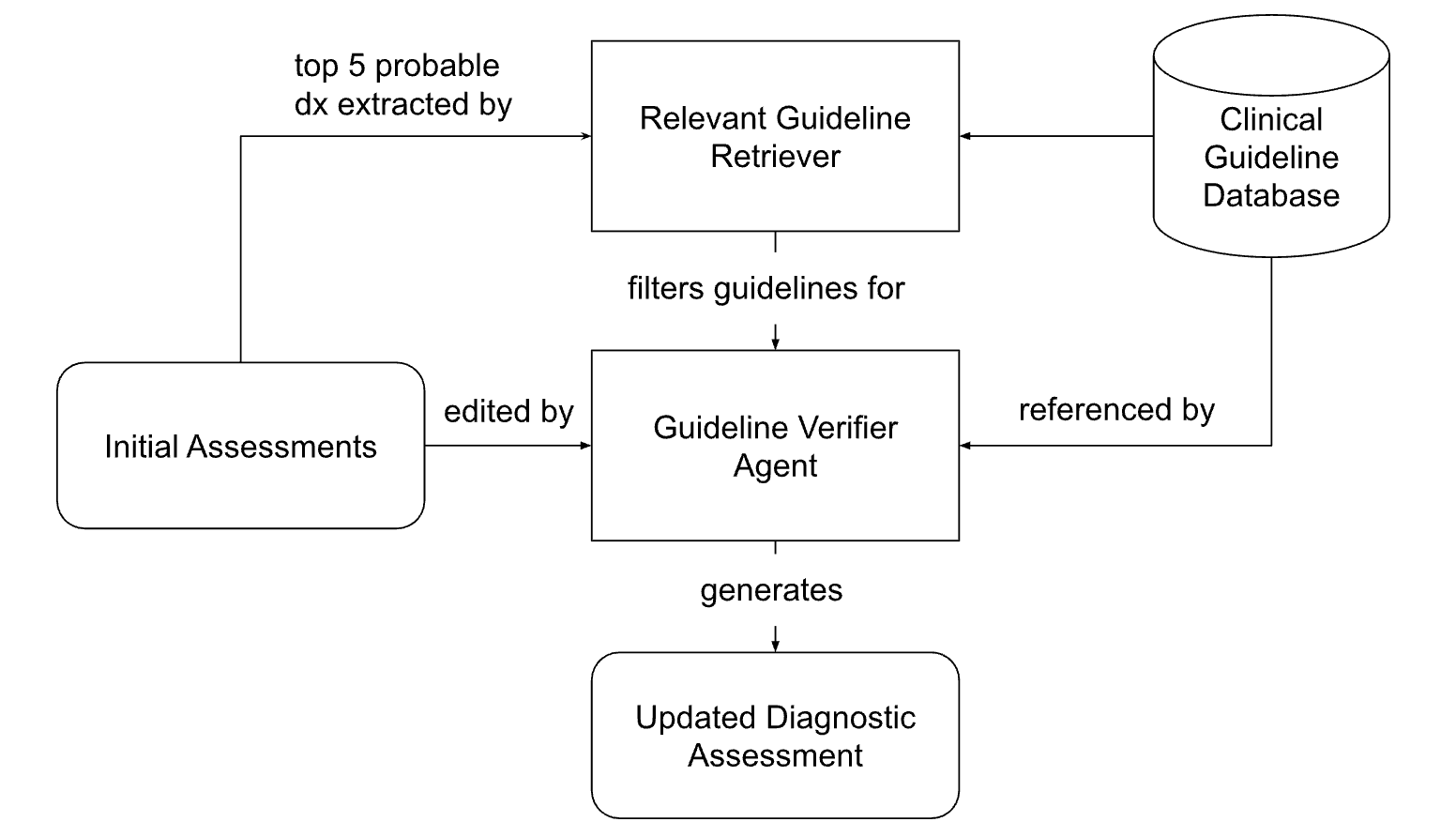}
    \caption{Guideline Verifier workflow. This figure shows how academically endorsed guidelines are recalled in real-time from verified sources and then utilized for modifying recommendations made by a system replying purely on knowledge embedded within a Large Language Model. }
    \label{fig:gv_arch}
\end{figure}

As an alternative strategy for determining urgency, we investigated the integration of diagnostic and treatment guidelines into a Retrieval Augmented Generation RAG workflow, which would constrain the language model to make decisions solely based on the retrieved guidelines. This approach explicitly restricts the system to conditions with well-established and comprehensive guidelines and offers a more interpretable decision-making process. For this investigation, we compiled a representative set of clinical guidelines.

The developed Guideline Verifier (GV) followed a structured pipeline. First, a GPT-4o model (acting as relevant guideline retriever) was given the top five differential diagnoses from the AI Triage assessments and retrieved corresponding clinical guidelines using LLM-based reasoning to capture conceptually similar conditions (rather than keyword matching). Then, the GV cross-referenced the recommended urgency level from the guidelines against the original assessment and adjusted the urgency if needed, defaulting to the most severe recommendation across the five (Figure~\ref{fig:gv_arch}).

The GV only ran if relevant guidelines were available for any of the top five diagnoses identified by the multi-agent system; otherwise, the original urgency status was retained. The GV was powered by the o3-mini model with RAG, selected for its strong reasoning ability in complex clinical tasks.

\section{Evaluations and Preliminary Results}

\subsection{Dataset and preprocessing}
We curated a dataset of deidentified real-world EHR records sourced from HealthVerity, a data provider. The dataset contained 21,779 deidentified records related to clinical encounters from 1,000 unique non-cancer patients, with each patient contributing multiple records. These encounters occurred between May 1, 2021, and April 30, 2024. Each record included routinely collected clinical data, such as patient demographics (age and gender), chief complaint, history of present illness, review of systems, past medical history, current medications, allergies, immunizations, social history, and family history.

To align the dataset with the study objective of evaluating the system’s triage capabilities, we filtered and structured the data around clinical scenarios most relevant to that task. First, we used GPT-4o-mini to classify the encounter cases into four distinct categories: \textit{Initial Encounter, Follow-up Visit, Routing Checkup, Unknown}.
The \textit{Initial Encounter} category included first visits for symptoms or health concerns that led to a new diagnosis or care plan. The \textit{Follow-Up Visit} category comprised subsequent visits intended to monitor progress, evaluate treatment, or check for complications. The \textit{Routine Checkup} category included preventive care visits, health maintenance evaluations, and chronic condition monitoring, often without the presence of symptoms. The \textit{Unknown} category included visits that lacked sufficient information for classification. In this work, we focused on \textit{Initial Encounter} cases as they cover the widest range of conditions while avoiding trivial information leakage from previous encounters.

Next, we used GPT-4o-mini to classify the retained cases based on the chief complaint and history of present illness into the following symptom categories: \textit{Pain-Related, Respiratory, Neurological, Gastrointestinal, Dermatological, Cardiovascular, Genitourinary, Musculoskeletal, Constitutional, and Psychological}. We excluded the \textit{Psychological} category, as mental health assessments typically require a fundamentally different triage framework involving specialized screening tools and crisis protocols. The \textit{Constitutional} category captured nonspecific systemic symptoms, such as chills or fever, that did not localize to a particular organ system.

Finally, we obtained a comparable number of cases from each of the nine retained symptom categories, resulting in 44–68 cases per category. The final dataset included 519 encounters, of whom 244 (47.01\%) were male and 275 (52.99\%) were female. These cases are then directly used in our Patient Simulator to instantiate and generate patient conversations with the AI Triage system.

\subsection{Evaluation}

We conducted an evaluation to gain insight into both the quality of Patient Simulator and the performance of AI Triage system across a series of domains of interest. Two clinician reviewers conducted an evaluation of the multi-agent output using a structured questionnaire containing 14 questions listed in \ref{sec:supp_eval} and an additional free text box to capture other noteworthy observations. The overall question architecture was such that the reviewers were asked whether they agreed with the system’s output (in a given aspect); therefore, ‘yes’ answers indicated positive agreement with the system’s output, while ‘no’ indicated an evaluation of the Patient Simulator's incoherence or of the triaging system’s output as incomplete / inappropriate / unsafe depending on the question. For each question, the ‘no’ was clearly defined in the answer options. On top of that, for each question reviewers were asked to add their reasoning if their answer differs from the model output.


The clinician reviewers were specialists in Internal Medicine or its related sub-disciplines, with a cumulative $\sim$50 years of experience. Reviewers had access to case vignettes, last prior-encounter notes, conversations between AI and patient agent, and all related questions and assessments. Reviewers completed their evaluations independently after reviewing those materials.


\subsection{Preliminary Results}

The questionnaire discussed above captures key aspects of the Patient Simulator and AI Triage. In this section, we report numbers where both physicians independently confirmed the model output. Starting with the Patient Simulator as a vital piece to make this simulation possible, it is consistent in 97.7\% of cases in communicating patient vignette symptoms to the AI Triage.

Further on, AI Triage asks precise and non-redundant questions in 81.7\%, has an appropriate tone in 99.6\%. 
After information gathering, the case summary in 99.2\% reflects chief complaints and key symptoms successfully.  Clinician reviewers found that the most likely diagnosis was among the top three differential diagnoses proposed by the agent in 95.4\% (495/519) of cases for one of the reviewers, and in 94.8\% (492/519) for the other.
Albeit some minor disagreements between the annotators regarding triage recommendation, at least one reviewer approved the final recommendation provided by AI Triage in majority of cases. The disagreements can be explained by the limited context available at the time of triage and by the different backgrounds of the physicians.
Interestingly, model agreement with physicians was more than agreement between themselves, cohen's kappa agreements 0.79, 0.74, and 0.72 between model vs physician 1, model vs physician 2, and physician 1 vs physician 2 respectively.
For the questionnaire-based evaluation, both clinician reviewers agreed with the system’s output in >90\% of cases for 7 out of 11 of the specific questions posed. 

\section{Discussion}
We have developed an evaluation framework for AI based triaging systems which stems from real patients encounters. Our methodology leveraged real-world EHR data to create simulated clinical encounters under a Patient Simulator. The Patient Simulator enables us to evaluate the interactions of conversational triage agents in a safe and offline environment. In parallel, we developed a multi-agent system framework AI Triage, consisting of several LLM-driven agents that work together to perform the \textit{initial assessment} and \textit{triage} roles of a healthcare professional.
\
Subsequently, we illustrate the evaluation pipeline applied to AI Triage against the Patient Simulator benchmark. The preliminary evaluation is encouraging, showing that the system’s output appears clinically appropriate in the majority of cases.
Moreover, we showed how an additional feature capable of retrieving external information, an agent for \textit{guideline verification}, can be incorporated in the system to further enhance the safety of the multi-agent system.
Our preliminary results offer insight into the reliability, strengths, and potential shortcomings of such a system in real-world applications through broad testing against a realistic Patient Simulator environment.

In our study, we first investigated how one may use real EHR to initialize a Patient Simulator using intermediary vignettes. Leveraging EHR allows us to simulate conversations with AI agents anchored on a real patient encounter, in order to evaluate triaging agents reasoning abilities.
However, being anchored in real data, our simulation suffers from traditional limitations of data inaccuracy, under-representation of specific groups or conditions. We have intentionally selected a balanced mixture of therapeutic areas, which may leave out less common symptoms or conditions.
Further our Patient Simulator was built off \textit{Initial Encounters} which we deemed were more complex for triaging purposes. However for returning patients, one might desire a more streamlined triaging process given the already established patient history, which is beyond the scope of our study.
Nevertheless, we believe our Patient Simulator lays the foundations for different applications, and can be extended for follow up patients or broader conditions.
\
The Patient Simulator leverages LLMs for conversational abilities. We have constructed a relatively simple annotation task to gauge the quality of the Patient Simulator which yielded satisfactory statistics, which guarantee coherence between experienced condition and the communicated symptoms.

Second, our study shows that multiple specialized agents overseen by a controller can emulate the process of making a clinical triage determination. This modular architecture, which represents a significant departure from monolithic LLM deployments, allowing interpretability, error localization, and task-specific optimization, has been the focus of recent developments in this field \citep{Mukherjee2024, johri2025evaluation, Schmidgall2024}. The integration of the GV agent introduces a second layer of validation grounded in clinical best practices, enhancing the safety and reliability of urgency assessments.
\
This approach brings forward several technical innovations to this field: the streamlined architecture is capable of real-time, coordinated interaction between multiple agents for comprehensive patient assessment; the dynamic data integration across various data types (labs, medications, symptoms) leverages patients’ historical data to provide contextually relevant and personalized triage; and the capacity for integrated complex step-by-step reasoning maximizes interpretability and transparency in the decision making process. Our initial observations are particularly promising in that they are supportive of the functionality of these features in an environment that, while simulated, was anchored on real-world data.

Despite these advancements, we acknowledge the limitations inherent in a simulation-based study. The limited context available at the time of conversational triage led to occasional discrepancies between physicians' assessments. Post-experiment interviews with annotators confirmed this, with both noting that some decisions were borderline and often came down to clinical judgment and individual preferences of the patient and physician.
Other limitations derive from the intrinsic features of our AI triage agent. Because the information exchange is entirely conversational, there may be gaps in content, particularly when patients provide vague or incomplete descriptions of their symptoms. However, this limitation is partially offset by the agents’ ability to retrieve structured medical data, such as lab results and medication histories, which help supplement patient-reported narratives.

\clearpage 

\bibliographystyle{apalike} 
\bibliography{references} 

\clearpage

\appendix 
\input{supplements}

\end{document}

%% file: supplements.tex

\section{Appendix}

\subsection{Patient Simulator Prompt}
\label{sec:supp_prompt}

This segment outlines the behavioral framework and communication guidelines for simulated patients participating in physician-patient interactions. The agent is instructed to present symptoms using natural, everyday language, avoid medical jargon, and respond authentically to physician prompts.
\begin{tcolorbox}[mygreybox]

    You're a patient experiencing a medical condition. Your background and symptoms are based on the following context:
    \texttt{<patient vignette>} 
    
    You are speaking with a physician who is working to diagnose your condition. Your goal is to respond to the physician as a real patient would.
    
    \textbf{Behaviors to Follow}
    \begin{enumerate}
        \item \textbf{Basic Knowledge \& Communication Style:}
        \begin{enumerate}[label=\alph*.]
            \item You have only basic medical knowledge and should describe your symptoms using everyday, non-professional language. For example, say “runny nose” instead of “sinus drainage” or “feeling tired” instead of “fatigue.”
            \item Offer symptoms broadly and conversationally, providing one or two symptoms at a time unless prompted for more details.
            \item Start the conversation with a brief description of your main concern (chief complaint), avoiding detailed or comprehensive symptom lists unless explicitly asked.
        \end{enumerate}
        \item \textbf{Avoid Professional Jargon:}
        \begin{enumerate}[label=\alph*.]
            \item Always use simple, layperson-friendly descriptions of symptoms.
            \item Do not use medical terms or technical phrases that a typical patient wouldn’t know or say.
        \end{enumerate}
        \item \textbf{Details Only When Asked:}
        \begin{enumerate}[label=\alph*.]
            \item Do not volunteer information about symptom severity, duration, or negative symptoms unless specifically questioned by the physician.
            \item Avoid providing detailed medical data (e.g., lab results, physical exam findings) that you wouldn’t reasonably know.
        \end{enumerate}
        \item \textbf{Common Sense Responses:}
        \begin{enumerate}[label=\alph*.]
            \item If a clear answer isn’t provided in the context, reply based on common sense. For example: If this is your first visit regarding these symptoms, assume you haven’t done related lab tests or scans, so answer “No.” For vaccinations, give reasonable responses based on your general profile. For instance, you likely received common vaccinations like the COVID-19 vaccine, so answer accordingly. Answer questions about symptoms not explicitly mentioned in your context with logical consistency (e.g., if not mentioned, assume you don’t have them).
        \end{enumerate}
        \item \textbf{Stay in Character:}
        \begin{enumerate}[label=\alph*.]
            \item Do not reference or imply the existence of a clinical vignette, scripts, or background information provided to you.
            \item Avoid saying anything like, “This information is not provided” or similar phrases. Respond as if you are unaware of any pre-existing script.
        \end{enumerate}
        \item \textbf{Parent/Guardian Role:}
        \begin{enumerate}[label=\alph*.]
            \item If the patient is a child or cannot represent themselves, speak as the patient’s parent, guardian, or caretaker based on the context.
        \end{enumerate}
    \end{enumerate}

\end{tcolorbox}

\newpage
\subsection{Evaluation Questionnaire}
\label{sec:supp_eval}

Evaluation rubric covering 14 key dimensions of the system:

\begin{tcolorbox}[mygreybox]

    \begin{enumerate}[label=\arabic*., leftmargin=*] 
    
        \item \textbf{Do the questions cover all important symptoms and related information?}
            \begin{itemize}[label=-, leftmargin=*, itemsep=0pt]
                \item Yes
                \item No, key information is missing
                    \begin{itemize}[label=, leftmargin=*, itemsep=0pt] 
                        \item \textit{\textbf{What key information was missing / not covered?} (Free text entry, displayed if "No, key information is missing")}
                    \end{itemize}
            \end{itemize}
    
        \item \textbf{Are questions precise and without redundancy?}
            \begin{itemize}[label=-, leftmargin=*, itemsep=0pt]
                \item Yes
                \item No, it contains irrelevant or repeated questions
                    \begin{itemize}[label=, leftmargin=*, itemsep=0pt]
                         \item \textit{\textbf{Please Explain} (Free text entry, displayed if "No, it contains irrelevant or repeated questions")}
                    \end{itemize}
            \end{itemize}
    
        \item \textbf{Does the doctor use an appropriate and empathetic tone with the patient during the conversation?}
            \begin{itemize}[label=-, leftmargin=*, itemsep=0pt]
                \item Yes
                \item No, Doctor's tone is sometimes inappropriate, rude, or not empathetic
                    \begin{itemize}[label=, leftmargin=*, itemsep=0pt]
                        \item \textit{\textbf{Please provide examples from the conversation} (Free text entry, displayed if "No, Doctor's tone is sometimes inappropriate, rude, or not empathetic")}
                    \end{itemize}
            \end{itemize}
    
        \item \textbf{Are the patient's answers consistent?}
            \begin{itemize}[label=-, leftmargin=*, itemsep=0pt]
                \item Yes
                \item No, the patient sometimes contradicts themself
                \item No, the patient sometimes contradicts the patient summary
                    \begin{itemize}[label=, leftmargin=*, itemsep=0pt]
                        \item \textit{\textbf{Please Explain} (Free text entry, displayed if "No, the patient sometimes contradicts themself" or "No, the patient sometimes contradicts the patient summary")}
                    \end{itemize}
            \end{itemize}
    
        \item \textbf{Does the list of requested historical EHR Data contain all necessary information for diagnosis?}
            \begin{itemize}[label=-, leftmargin=*, itemsep=0pt]
                \item Yes
                \item No
                \item N/A (if list is empty)
                    \begin{itemize}[label=, leftmargin=*, itemsep=0pt]
                        \item \textit{\textbf{What key information was missing / not covered?} (Free text entry, displayed if "No")}
                    \end{itemize}
            \end{itemize}
    
        \item \textbf{Does the Case Summary capture the chief complaint and key information necessary (positive signs and important negative signs) for diagnosis?}
            \begin{itemize}[label=-, leftmargin=*, itemsep=0pt]
                \item Yes
                \item No
                    \begin{itemize}[label=, leftmargin=*, itemsep=0pt]
                        \item \textit{\textbf{What important information was missing?} (Free text entry, displayed if "No")}
                    \end{itemize}
            \end{itemize}
    
        \item \textbf{Does the Laboratory Assessment make accurate inferences about the patient’s condition?}
            \begin{itemize}[label=-, leftmargin=*, itemsep=0pt]
                \item Yes
                \item No, inference about some labs are incorrect
                \item N/A (if no labs are available)
                    \begin{itemize}[label=, leftmargin=*, itemsep=0pt]
                        \item \textit{\textbf{Please elaborate} (Free text entry, displayed if "No, inference about some labs are incorrect")}
                    \end{itemize}
            \end{itemize}
    
        \item \textbf{Does the Medication Assessment make accurate inferences about the patient's condition?}
            \begin{itemize}[label=-, leftmargin=*, itemsep=0pt]
                \item Yes
                \item No, inference about some meds are incorrect
                \item N/A (if no meds are available)
                    \begin{itemize}[label=, leftmargin=*, itemsep=0pt]
                        \item \textit{\textbf{Please elaborate} (Free text entry, displayed if "No, inference about some meds are incorrect")}
                    \end{itemize}
            \end{itemize}
    
        \item \textbf{Does the Overall Assessment draw accurate conclusions based on the conversation and provided EHR?}
            \begin{itemize}[label=-, leftmargin=*, itemsep=0pt]
                \item Yes
                \item No, the conclusion is inaccurate
                    \begin{itemize}[label=, leftmargin=*, itemsep=0pt]
                        \item \textit{\textbf{Please elaborate} (Free text entry, displayed if "No, the conclusion is inaccurate")}
                    \end{itemize}
            \end{itemize}
    
        \item \textbf{Is the reasoning in the Urgency Assessment correct?}
            \begin{itemize}[label=-, leftmargin=*, itemsep=0pt]
                \item Yes
                \item No
                    \begin{itemize}[label=, leftmargin=*, itemsep=0pt]
                        \item \textit{\textbf{Please explain what is incorrect, if anything} (Free text entry, displayed if "Yes" or "No")}
                    \end{itemize}
            \end{itemize}
    
        \item \textbf{What is the correct Urgency Status? (Select multiple if applicable)}
            \begin{itemize}[label=-, leftmargin=*, itemsep=0pt]
                \item Self-care
                \item Follow up with PCP
                \item Urgent care / emergency
                    \begin{itemize}[label=, leftmargin=*, itemsep=0pt]
                        \item \textit{\textbf{If your selection does not match the provided 'Urgency Status', please give your reasoning} (Free text entry, displayed if "Self-care", "Follow up with PCP", or "Urgent care / emergency" is selected and differs from system)} 
                    \end{itemize}
            \end{itemize}
    
        \item \textbf{Does the 'when to escalate?' help patient understanding urgency and when to take additional steps?}
            \begin{itemize}[label=-, leftmargin=*, itemsep=0pt]
                \item Yes
                \item No, it contains harmful or misleading information
                    \begin{itemize}[label=, leftmargin=*, itemsep=0pt]
                        \item \textit{\textbf{Please elaborate} (Free text entry, displayed if "No, it contains harmful or misleading information")}
                    \end{itemize}
            \end{itemize}
    
        \item \textbf{Are the 'Care Recommendations' helpful based on the patient condition?}
            \begin{itemize}[label=-, leftmargin=*, itemsep=0pt]
                \item Yes
                \item No, it contains harmful or misleading information
                    \begin{itemize}[label=, leftmargin=*, itemsep=0pt]
                        \item \textit{\textbf{Please elaborate} (Free text entry, displayed if "No, it contains harmful or misleading information")}
                    \end{itemize}
            \end{itemize}
    
        \item \textbf{Please select the most probable diagnosis (choose multiple if more than 1 are equally likely given the available information).}
            \begin{itemize}[label=-, leftmargin=*, itemsep=0pt]
                \item Diagnosis 1
                \item Diagnosis 2
                \item Diagnosis 3
                \item Diagnosis 4
                \item Diagnosis 5
                \item Other (Not Listed)
                    \begin{itemize}[label=, leftmargin=*, itemsep=0pt]
                        \item \textit{\textbf{Enter most probable diagnosis here.} If multiple diagnosis missing, provide comma separated list - e.g., 'Flu, Covid'. (Free text entry)}
                        \item \textit{\textbf{Provide any comments on diagnosis here:} (Free text entry)}
                    \end{itemize}
            \end{itemize}
    
    \end{enumerate}
    
    \vspace{1em} 
    
    \noindent\textbf{General / additional comments:}
    \begin{itemize}[label=, leftmargin=*, itemsep=0pt]
        \item \textit{(Free text entry)}
    \end{itemize}
\end{tcolorbox}